\documentclass[table]{stellaredge}

\usepackage{times}          % serif body font (Times); comment out for Latin Modern
\usepackage[utf8]{inputenc}
\usepackage{amsmath,amssymb,amsfonts}
\usepackage{graphicx}
\usepackage{booktabs}
\usepackage{xspace}
\usepackage{wrapfig}

\crefname{figure}{Fig.}{Figs.}\Crefname{figure}{Fig.}{Figs.}
\crefname{table}{Tab.}{Tabs.}\Crefname{table}{Tab.}{Tabs.}
\crefname{section}{Sec.}{Secs.}

\title{StellaVLA: In-Context Structured Demonstration for Generalizable Vision-Language-Action Models}

\affiliation{StellarEdge AI Technical Team}

\date{August 1, 2026}
\metadata{\href{https://stelledge.com/blog/stellavla}{https://stelledge.com/blog/stellavla}}
\abstract{%
Vision-Language-Action (VLA) models can follow instructions and manipulate objects, but their performance often collapses out of distribution (OOD), when the scene, viewpoint, or object differs from training. Adapting to each new situation typically requires collecting more data and fine-tuning. We present \textbf{StellaVLA}, a framework that instead adapts at test time by conditioning on a single retrieved demonstration. The key idea is to move beyond imitating \emph{what} an expert did and instead convey \emph{why}: an automated offline pipeline converts each raw trajectory into a \emph{structured demonstration}, \textit{e.g.}, a task plan, sub-goal descriptions, and verbalized 3D motion, at zero human-annotation cost. Provided as in-context guidance, this structured demonstration lets the policy reason about the task rather than mimic a pixel trajectory, which also makes it transferable across embodiments (real-robot, human-hand, or XR demonstrations).
A parallel dual-training design internalizes this reasoning during training through a joint action-and-language objective, while inference uses the action expert alone, preserving real-time, high-frequency control with no added latency. On the VLA-Arena leaderboard\footnote{https://vla-arena.github.io/\#leaderboard} (Aug 1, 2026), StellaVLA ranks \textbf{first} with an overall score of 0.63, versus 0.44 and 0.22 for the strong prior models ($\pi_{0.5}$ and LingBot-VLA), and it further leads on LIBERO with 98.8\% average success rate and LIBERO-Plus with 85.1\% success rate. Our real-robot benchmark demonstrates that StellaVLA can use both human/robot demos and human-to-robot (XR) demos as in-context structured demonstration to help VLA model adapt to OOD tasks. 
}

\begin{document}
\maketitle

\begin{figure}[t]
\centering
\includegraphics[width=\linewidth]{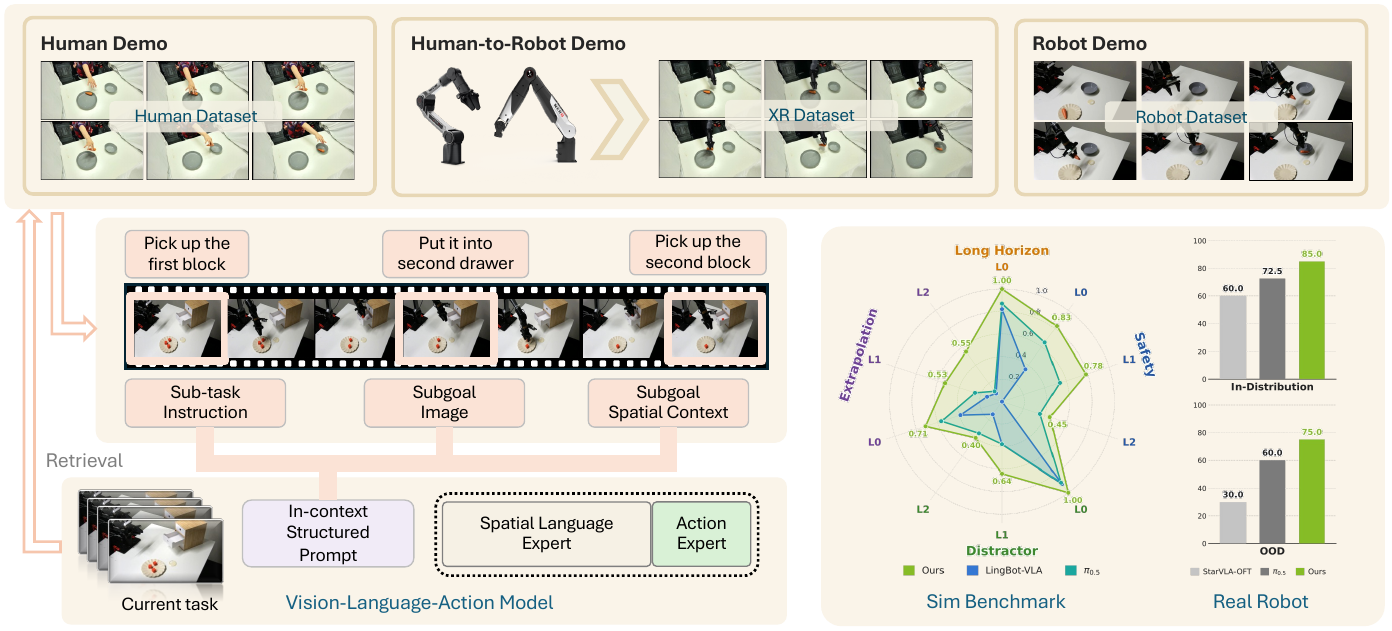}
\caption{Overview of \textbf{StellaVLA}, which conditions a VLA policy on
in-context structured demonstrations. We convert diverse human and robot
demonstrations into structured examples. Each example comprises high-level
\emph{semantic} rationales (sub-goals) and fine-grained \emph{kinematic}
rationales (movements), which are retrieved to prompt a vision-language-action
model for manipulation tasks. During training, an auxiliary spatial-language
expert supervises these rationales so the policy grounds its actions in the
underlying reasoning rather than only imitating the observed motions. At
inference this expert is removed and only the action expert is used,
preserving real-time control. The resulting model generalizes more robustly to
unseen tasks and environments in both simulation and real-world experiments.}
\label{figs:start}
\end{figure}

\section{Introduction}

Vision-Language-Action (VLA) models have become a prominent paradigm for robotic manipulation~\cite{openvla, pi0}. Built on pretrained Vision-Language Models (VLMs), they translate visual and textual inputs into physical actions. In practice, however, their performance degrades sharply out of distribution (OOD), when the scene, viewpoint, or object differs from training~\cite{fang2025intention, zhou2025libero, xu2026affordance}, and recovering typically requires collecting new data and fine-tuning.
In-Context Imitation Learning (ICIL) offers a way to adapt at test time without weight updates~\cite{fu2025icrt, retrievalvla}. By appending retrieved expert trajectories, \textit{e.g.}, sequences of observations and low-level actions, as a contextual prefix, ICIL gives the policy a non-parametric memory it can imitate on the fly. But existing frameworks use these trajectories only as raw observations and continuous actions, which encourages surface-level imitation: the policy sees \emph{what} the expert did without the \emph{why}. Lacking that structure, it often treats the demonstration as noise and falls back on its pretrained priors~\cite{xu2026affordance}, which is a behavioral inertia that keeps it from generalizing to out-of-distribution tasks.

Reasoning provides the missing structure. Foundation models generalize better when the intermediate reasoning is made explicit rather than left implicit~\cite{cotvla,acotvla, zawalski2024robotic, huang2026thinkact}: decomposing a task into logical steps exposes the \emph{why} behind each action. The question we address is how to surface this reasoning inside retrieved demonstrations, so that a VLA policy imitates the expert's reasoning rather than only its motions.
Motivated by this idea, we propose \textbf{StellaVLA} (Figure~\ref{figs:start}), which advances ICIL by turning demonstrations into structured, reasoning-augmented context. An automated offline pipeline converts raw expert trajectories into this form at zero human-annotation cost: following the ``language-as-action'' paradigm~\cite{lin2026la4vla, steerablevla}, a VLM segments long-horizon trajectories and generates hierarchical rationales, high-level \emph{semantic} rationales (sub-goals) and fine-grained \emph{kinematic} rationales (movements), which augment each demonstration alongside its original observations and actions. Applied across sources, this yields a unified cross-embodiment demonstration pool.

Unlike prior ICIL methods that rely on prompting alone~\cite{zhang2026retrieval, jangra}, StellaVLA couples in-context prompting with explicit supervision through a parallel dual-training design. During training, the model conditions on the retrieved rationale-augmented demonstrations and is jointly optimized to predict low-level actions and to articulate the corresponding rationales, internalizing both \emph{what} to do and \emph{why}. The language supervision grounds actions in their semantic intent, while the retrieved rationales act as a reusable template for how to plan in similar situations. At inference, the spatial-language expert is removed entirely: the policy uses only the action expert, guided by the KV-cached demonstration prefix, giving real-time high-frequency control with no added latency while still benefiting from the task structure learned during training.

Our main contributions are:
\begin{itemize}
    \item We propose \textbf{StellaVLA}, a retrieval-augmented VLA framework that shifts ICIL from imitating actions to imitating the reasoning behind them, reducing behavioral inertia under OOD conditions.
    \item A parallel dual-training design, fed by a zero-human-cost offline extraction pipeline, that internalizes expert reasoning during training and strips the spatial-language expert at inference, leaving the control loop free of autoregressive decoding overhead.
    \item Extensive simulation and real-robot manipulation experiments show that StellaVLA ranks \textbf{first} on the VLA-Arena leaderboard (Aug 1, 2026) with an overall score of \textbf{0.63}, while achieving \textbf{98.8\%} average success in LIBERO and leading in LIBERO-Plus and our real-robot benchmark.
\end{itemize}

\section{Related Work}

Vision-language-action (VLA) models adapt pretrained vision-language backbones for robotic control. Actions can be generated as discretized tokens~\citep{openvla}, or predicted through regression, flow-matching, and frequency-domain action heads~\citep{openvlaoft,pi0,pi05,pi0fast,hemotion}. Recent work has further improved computational efficiency~\citep{pei2026action,xu2026vla,feng2026see,xie2026towards} or augmented policy representations with geometric, depth, and world-model priors~\citep{groot,smolvla,nora,starvla,cogvla,evodepth,dreamvla}. In contrast, StellaVLA retains a standard VLM backbone with a continuous-action expert and focuses on a complementary question: how expert demonstrations should be represented and exploited as in-context supervision.

\subsection{Language as an Action Representation}
Language-as-action methods represent robot behavior using the vocabulary of
pretrained VLMs, providing a semantic interface between high-level reasoning
and continuous control. Prior work verbalizes low-level actions, jointly
models language and action tokens, or represents action hierarchies and
subtasks in language~\citep{lap,actionsaslang,la4vla,rth2024arxiv}. By
expressing behavior in the VLM's semantic space, these approaches retain an
interpretable correspondence between task intent and robot motion.

StellaVLA uses language as a shared representation for both retrieved context
and training supervision. Each demonstration is organized into subgoals that
pair semantic descriptions and keyframes with language-rendered robot states
and 2D/3D motions, yielding a structured semantic state-action sequence.
During training, a language expert predicts the current subtask and a
language rendering of the same action chunk predicted by the continuous
action expert. This shared interface connects demonstration-level procedural
structure with step-level control, while the language expert is removed at
inference. This differs from latent-action approaches, which encode behavior
into learned latent variables outside the VLM's native semantic
space~\citep{lapa,bu2025univla}.

\subsection{Test-Time Adaptation and Context-Conditioned VLAs}
Robot policies can adapt at deployment either by updating model parameters
or by conditioning a fixed policy on additional experience. Test-time
training methods update fast weights, latent prompts, or memory using
deployment data~\citep{ttt,robottt,tttvla,wamttt}. In contrast,
context-conditioned approaches leave the policy parameters unchanged and
adapt behavior through demonstrations, retrieved experience, or execution
history~\citep{behaviorprompt,retrievalvla,locoformer,memoryvla}. Recent work
has further explored in-context imitation and retrieval-based adaptation for
VLAs~\citep{fu2025icrt,sridhar2025ricl,jang2026ravla}.

The supplied context, however, can play substantially different roles.
$\pi_{0.7}$ conditions on subtask instructions, subgoal images, episode
metadata, and control modes to specify an execution strategy~\citep{pi07},
while Qwen-RobotManip uses recent observation--state--action chunks to adapt
to the dynamics of the current episode~\citep{qwenrobotmanip}. StellaVLA
instead retrieves a prior demonstration and converts it into a structured
procedural context composed of segment-level subtasks and grounded 2D/3D
motion. The retrieved episode therefore serves as an explicit task
demonstration rather than merely additional execution history, enabling
adaptation without parameter updates.

\subsection{Human and Cross-Embodiment Demonstrations}
Human videos and heterogeneous robot data provide broader behavioral coverage
than single-platform demonstrations, but differ in both embodiment and action
space. Prior work addresses this mismatch through shared latent actions,
cross-embodiment motion representations, world models, video conditioning,
retargeting, or canonicalized action spaces~\citep{univla,motus,dreamdojo,
vivla,wang2026x,qwenrobotmanip,hui2026seeing}.

StellaVLA instead addresses cross-embodiment demonstrations at the
representation level. Real-robot, human-hand, and XR-retargeted
demonstrations are converted into a common structured representation of
semantic subtasks and grounded motion before retrieval. Crucially,
off-embodiment demonstrations are used only as context, and executable actions
are always predicted and supervised in the target robot's native control
space. This separation allows heterogeneous demonstrations to provide
procedural guidance without requiring their source action spaces to be
directly aligned with the target policy.

\section{Methodology}
\begin{figure}[t]
\centering
\includegraphics[width=\linewidth]{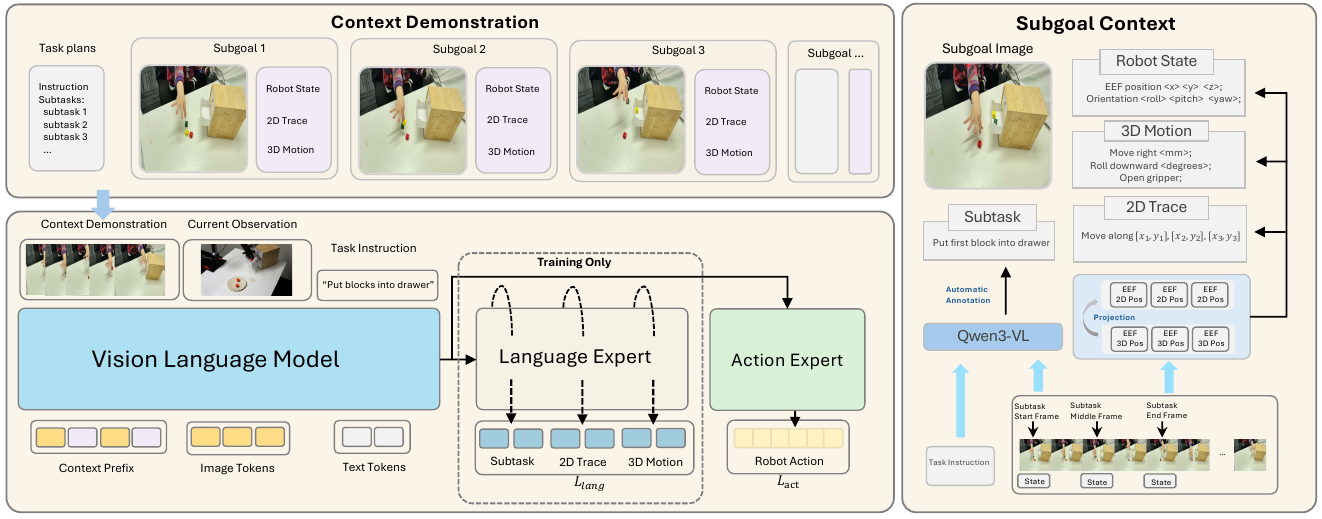}
\caption{\textbf{Overview of StellaVLA.} \emph{Top left:} a retrieved context demonstration is represented as a task plan and a sequence of subgoals, each containing a keyframe, robot state, 2D trace, and 3D motion. \emph{Right:} these annotations are generated automatically offline from VLM reasoning and robot trajectories, without human labeling. \emph{Bottom left:} the vision-language model encodes the demonstration together with the current observation and instruction into a shared representation, which is consumed by an action expert for action prediction and a spatial-language expert for auxiliary supervision during training only. The spatial-language expert is removed at inference, requiring only a single forward pass.}
\label{fig:overview}
\end{figure}

In this section, we detail the \textbf{StellaVLA} framework, illustrated in Figure~\ref{fig:overview}, which advances ICIL with structured contextual rationales to improve VLA generalization.
We first present an automated offline pipeline that converts raw trajectories into rationale-augmented segments (Sec.~\ref{sec:offline_extraction}), then introduce a parallel dual-training paradigm that uses these trajectories as both explicit supervision and in-context prompts (Sec.~\ref{sec:dual_training}), and finally describe an asymmetric inference strategy that leaves the control loop free of autoregressive decoding overhead (Sec.~\ref{sec:inference}).

% ---------------------------------------------------------------------
% 3.1 Offline Structured Rationale Distillation
% ---------------------------------------------------------------------
\subsection{Offline Structured Context Extraction}
\label{sec:offline_extraction}

\paragraph{Raw trajectory formulation.}
An expert demonstration is recorded as $\tau = \{(o_t, a_t)\}_{t=1}^T$, where $o_t$ denotes sensory input (\textit{e.g.}, multi-view RGB and proprioception) and $a_t \in \mathbb{R}^{d_a}$ is the continuous control command (\textit{e.g.}, end-effector pose and gripper state).
Such trajectories may come from diverse embodiments and collection paradigms, including real-robot teleoperation, human hand tracking, or XR-retargeted demonstrations~\cite{wang2026x}. 
Although they encapsulate successful executions, the raw numerical actions lack explicit semantic rationale, which hinders cross-embodiment reasoning.

\paragraph{Semantic segmentation via causal deduction.}
To bridge the semantic gap without incurring prohibitive human annotation costs, we introduce an automated offline pipeline (Figure~\ref{fig:overview}, right) that deduces the expert's underlying thought process. 
The core motivation is to infer the ``cause'' (the expert's underlying structured rationale) from the ``effect'' (the observed physical execution).
Given a raw trajectory $\tau$ and its corresponding high-level task instruction $\mathcal{I}$, we utilize a powerful off-the-shelf Vision-Language Model (e.g., Qwen3-VL) to explicitly decompose the continuous trajectory into $K$ discrete, semantically meaningful segments. 
For each segment $k \in \{1, \dots, K\}$ spanning from time step $t_{\mathrm{start}}^{(k)}$ to $t_{\mathrm{end}}^{(k)}$, the VLM analyzes the visual changes to identify the high-level sub-goal achieved. This process captures the true decision-making process that governs the expert's behavior.

\paragraph{Kinematic verbalization via language-as-action.}
Once the trajectory is segmented, we adopt a ``language-as-action'' paradigm that deterministically translates physical actions within each segment into structured text. 
Because VLAs are built upon VLMs with a strong affinity for semantic language\cite{actionsaslang,zhang2026revisiting}, these rationales can be processed with the original text vocabulary, without introducing specialized action tokens. 

Ultimately, this two-tier process generates a \textbf{Structured Context} $l_k$ for each segment that explicitly articulates the underlying logic through a hierarchical representation:
\begin{itemize}
    \item \textbf{Semantic Rationale (Sub-goal Description):} Captures the high-level semantic objective achieved in this segment (e.g., \textit{``Reach for the handle of the blue mug''}), automatically identified by the VLM based on visual observations.
    \item \textbf{Kinematic Rationale (Movement Description):} Verbalizes fine-grained motion as a \textbf{3D movement} in the workspace (e.g., \textit{``Move the end-effector by $\Delta x=+0.05, \Delta y=-0.02, \Delta z=+0.10$ and close gripper''}) and a \textbf{2D movement} obtained by projecting the 3D trajectory onto the camera plane with known intrinsics/extrinsics. This dual form encourages the VLM to reason about spatial logic and visual grounding via text tokens, rather than emitting raw floating-point actions.
\end{itemize}

Both kinematic fields are produced by a single deterministic verbaliser
$\Phi$, which maps any contiguous span of actions to its 3D displacement in
the workspace and to the 2D projection of that displacement onto the image
plane. Applying $\Phi$ over a whole segment yields the movement description
carried inside $l_k$; we reuse the \emph{same} operator at the much shorter
scale of an action chunk in Sec.~\ref{sec:dual_training}, which is what
places the retrieved demonstration and the policy's own prediction in one
common vocabulary. The sub-goal description, by contrast, is attached to
every frame: each step $t$ carries the subtask $s_t$ that the expert is
executing in the observation $o_t$.

\paragraph{Rationale-augmented demonstration pool.}
Through this process, the original raw trajectory $\tau$ is transformed into a rationale-augmented trajectory $\tau_{\mathrm{rat}} = \{(o_t, a_t, s_t, l_k)\}_{t=1}^T$, where each time step $t$ is now grounded not only by its corresponding observation and action, but also by its own subtask label $s_t$ and by the explicit \emph{segment-level} structured rationale $l_k$ of the segment it belongs to.
By processing all available expert demonstrations through this pipeline, we construct a comprehensive demonstration pool $\mathcal{D}_{\mathrm{pool}} = \{\tau_{\mathrm{rat}}^{(i)}\}_{i=1}^N$.

% ---------------------------------------------------------------------
% 3.2 Parallel Dual-Training Paradigm
% ---------------------------------------------------------------------
\subsection{Parallel Dual-Training Paradigm}
\label{sec:dual_training}

\paragraph{Retrieval-augmented prompt formulation.}
During training, the learning process is formulated as a leave-one-out retrieval task.
To train the policy to imitate a specific target trajectory $\tau_{\mathrm{tgt}}$ sampled from $\mathcal{D}_{\mathrm{pool}}$, the system queries the remaining pool $\mathcal{D}_{\mathrm{pool}} \setminus \{\tau_{\mathrm{tgt}}\}$ and retrieves the top-$1$ rationale-augmented expert trajectory, ranked by cosine similarity between the language embeddings of the task instructions. Retrieval is therefore purely linguistic, which is what lets a demonstration recorded on another embodiment be retrieved for a robot episode.
The retrieved trajectory is formatted into a contextual prefix prompt $\mathcal{P}_{\mathrm{demo}}$. The final input to the VLA model at time step $t$ is constructed by concatenating the retrieved demonstration, the current task instruction $\mathcal{I}$, and the current observation $o_t$:
\begin{equation}
    x_t = \bigl[ \mathcal{P}_{\mathrm{demo}} \oplus \mathcal{I} \oplus o_t \bigr]
\end{equation}
Crucially, because $\mathcal{P}_{\mathrm{demo}}$ contains the rich $l_k$ descriptions from the experts, it acts as an implicit in-context learning (ICL) signal, providing the model with a cognitive template of how to ``think'' and plan in similar scenarios.

\paragraph{Parallel experts architecture.}
The core of our VLA policy is a unified Vision-Language Model backbone $f_\theta$ equipped with two parallel experts (Figure~\ref{fig:overview}, bottom left). Given the input $x_t$, the backbone processes the multimodal sequence into a shared latent representation $h_t = f_\theta(x_t)$. This single latent vector is then simultaneously fed into:
\begin{enumerate}
    \item An \textbf{action expert}, a lightweight MLP head that regresses the
          continuous action chunk
          $\hat{A}_t = (\hat{a}_t, \dots, \hat{a}_{t+H-1})$ spanning the next
          $H$ control steps.
    \item A \textbf{spatial-language expert}, the native autoregressive LM
          head, that emits the rationale
          $\hat{c}_t = (\hat{s}_t, \hat{m}_t)$ of the current step: the
          subtask $\hat{s}_t$ being executed in $o_t$, and the verbalised 3D
          and 2D movement $\hat{m}_t$ of that very same action chunk.
\end{enumerate}
It is vital to note that these two experts operate strictly in parallel. The action prediction does not causally depend on the generated text tokens at time step $t$. They share the representation $h_t$ and nothing else.

\paragraph{Spatial language supervision.}
While the retrieved demonstration provides segment-level context,
spatial-language supervision is defined at each time step as
\begin{equation}
    c_t = \bigl(s_t,\; \Phi(A_t)\bigr),
    \label{eq:chunktarget}
\end{equation}
where $s_t$ is the subtask label for $o_t$, $A_t = (a_t, \dots,
a_{t+H-1})$ is the ground-truth action chunk, and $\Phi(A_t)$ verbalises its
3D and 2D movement. The subtask label provides semantic supervision for the
current task stage; the 2D movement grounds the prediction in $o_t$, while
the 3D movement provides kinematic supervision in the workspace.

\paragraph{Dual rationale objective.}
During training, the model is optimized to minimize a joint objective:
\begin{equation}
    \mathcal{L} = \mathcal{L}_{\mathrm{act}}(\hat{A}_t, A_t) + \lambda \mathcal{L}_{\mathrm{lang}}(\hat{c}_t, c_t)
\end{equation}
where $\mathcal{L}_{\mathrm{act}}$ is the regression loss for the continuous control commands (e.g., $L_1$ loss), $\mathcal{L}_{\mathrm{lang}}$ is the standard autoregressive cross-entropy loss for the structured rationale text generation, and $\lambda$ is a balancing coefficient.

The two experts provide complementary supervision for the shared
representation $h_t$. Since $\Phi$ is deterministic, $A_t$ and
$\Phi(A_t)$ describe the same motion in continuous and linguistic forms.
Together, the explicit semantic, grounding, and kinematic supervision in
$c_t$ and the contextual guidance from $\mathcal{P}_{\mathrm{demo}}$
encourage $h_t$ to connect the demonstrated task structure and current
observation with the robot motion required at time $t$.

% ---------------------------------------------------------------------
% 3.3 Asymmetric Inference and Caching
% ---------------------------------------------------------------------
\subsection{Asymmetric Inference and Caching}
\label{sec:inference}

\paragraph{Action-only execution.}
A fundamental limitation of standard Embodied CoT paradigms is the severe latency introduced by autoregressive text generation during inference\cite{cotvla, zawalski2024robotic,lee2025molmoact, huang2026thinkact}, which often breaks the strict timing requirements of high-frequency robotic control. Our parallel architecture elegantly resolves this bottleneck. Because the action expert and the spatial-language expert read from $h_t$ independently, and the profound physical ``mental model'' has already been forged into the backbone weights during training via $\mathcal{L}_{\mathrm{lang}}$, the language read-out becomes strictly optional at deployment.

During inference, the autoregressive spatial-language expert is entirely stripped away. The policy executes a single forward pass through the backbone and the lightweight MLP action expert to output the action chunk $\hat{A}_t$. This asymmetric strategy decouples the acquisition of reasoning (paid for during training) from its execution, so continuous control incurs no autoregressive decoding overhead.

\paragraph{Demonstration prefix caching.}
To further optimize inference efficiency, we exploit the static nature of the retrieved context\cite{xu2026vla}. For a given novel task, the retrieved rationale-rich demonstration prefix $\mathcal{P}_{\mathrm{demo}}$ is pinned at the beginning of the rollout and remains immutable for the duration of the episode. Consequently, the key-value (KV) cache for the entire prefix is computed exactly once at $t=1$. For all subsequent time steps, the policy only needs to forward the live suffix (the current observation $o_t$), drastically reducing the computational footprint. This caching mechanism ensures that carrying a rich, multi-step rationale demonstration incurs virtually zero marginal cost during the high-frequency control loop.

\section{Experiments}

We evaluate four questions: (i) whether structured demonstrations improve
in-distribution performance and generalization under task and visual shifts,
(ii) whether the policy uses the retrieved demonstration and which parts of
it carry transferable information, (iii) how spatial-language supervision
affects the learned representation, and (iv) whether the resulting policy
can use cross-embodiment context without sacrificing deployment efficiency.

\subsection{Experimental Setup}
\noindent \textbf{Model and Training.}
StellaVLA couples a Qwen3-VL-4B backbone~\citep{qwen3vl} with an
OpenVLA-OFT-style MLP action expert~\citep{openvlaoft}. At each step, the
model receives third-person and wrist RGB observations, a language rendering
of the robot state, the task instruction, and one retrieved same-task
structured demonstration. The action expert regresses an action chunk under
an $L_1$ objective. In parallel, the spatial-language expert is trained by
cross-entropy (weight $\lambda=0.3$) to predict the current subtask and the
2D/3D description of the same action chunk. The subtask, 2D movement, and
3D movement provide semantic, visual-grounding, and kinematic supervision,
respectively. The spatial-language expert is not decoded at inference.

All models are fully fine-tuned from Qwen3-VL-4B-Instruct for $30$k steps
with a global batch size of $128$. Context-demonstration dropout is $0.0$
when a same-task demonstration is always available and $0.5$ otherwise.
Appendix~A gives optimization, retrieval, and platform-specific details.

\noindent \textbf{Protocol and Matched Control.}
We report task success rate under each benchmark's official protocol.
\emph{StarVLA-OFT}~\citep{starvla} is trained with the same backbone, action
expert, and data as StellaVLA, but receives neither a retrieved demonstration
nor spatial-language supervision. It therefore measures the gain from the
complete StellaVLA design rather than either component alone. To isolate the
role of context, Table~\ref{tab:abl_content} instead intervenes on a fixed
StellaVLA checkpoint by supplying the correct, no, or a wrong-task
demonstration at evaluation.

\subsection{Generalization in Simulation}
\begin{figure}[!t]
\centering
\includegraphics[width=\linewidth]{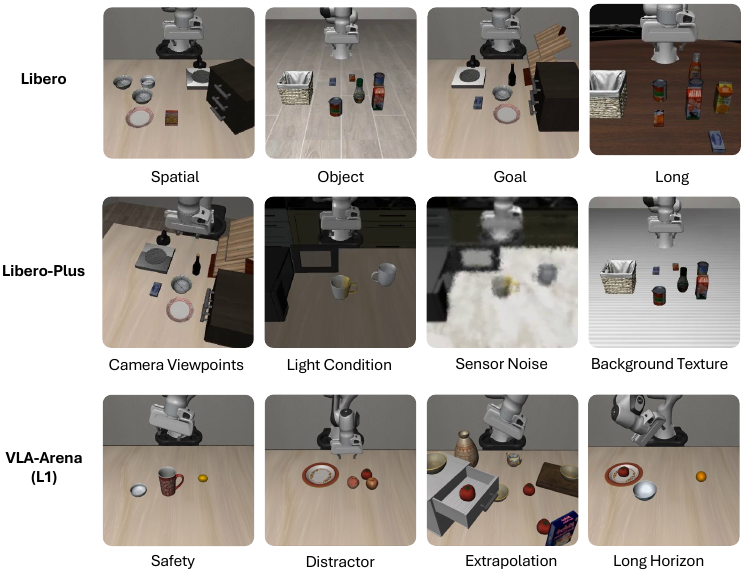}
\caption{\textbf{The three simulation benchmarks.} \emph{Top}: the four standard LIBERO suites. \emph{Middle}: LIBERO-Plus perturbation axes, e.g., camera viewpoint, lighting, sensor noise and background texture. \emph{Bottom}: VLA-Arena, which changes the task, adding safety constraints, distractors, extrapolation to unseen objects, and long-horizon composition.}
\label{fig:simulation_benchmarks}
\end{figure}

We first evaluate StellaVLA on three simulation benchmarks, as shown in Figure~\ref{fig:simulation_benchmarks}, probing in-distribution competence, task-level generalization, and zero-shot
robustness. 

\begin{table}[t]
\centering
\small
\setlength{\tabcolsep}{16pt}
\begin{tabular}{lccccc}
\toprule
Method & Spatial & Object & Goal & Long & Avg. \\
\midrule
MemoryVLA~\citep{memoryvla} & 98.4 & 98.4 & 96.4 & 93.4 & 96.7 \\
ACoT-VLA~\citep{acotvla} & 99.4 & \textbf{99.6} & 98.8 & 96.0 & 98.5 \\
AVA-VLA~\citep{avavla} & 99.2 & \textbf{99.6} & 97.9 & 96.2 & 98.2 \\
StarVLA-OFT & 97.8 & 98.6 & 96.2 & 93.8 & 96.6 \\
CogVLA~\citep{cogvla} & 98.6 & 98.8 & 96.6 & 95.4 & 97.4 \\
Retrieval-VLA~\citep{retrievalvla} & 97.4 & 98.8 & 96.3 & 89.5 & 95.5 \\
DreamVLA~\citep{dreamvla} & 97.5 & 94.0 & 89.5 & 89.5 & 92.6 \\
\midrule
\textbf{StellaVLA (Ours)} & \textbf{99.6} & 99.0 & \textbf{99.6} & \textbf{96.8} & \textbf{98.8} \\
\bottomrule
\end{tabular}
\caption{In-distribution success rate (\%) on the standard LIBERO
benchmark~\citep{libero}. Each suite is evaluated over $500$ rollouts;
baseline numbers are as reported in the original papers, except our matched
demonstration-free control StarVLA-OFT~\citep{starvla}. Best per column in
\textbf{bold}.}
\label{tab:libero}
\end{table}

\subsubsection{In-Distribution Validation in LIBERO.}
StellaVLA reaches $98.8\%$ average success on LIBERO
(Table~\ref{tab:libero}), showing that structured context does not compromise
in-distribution performance. Relative to StarVLA-OFT, the gain is small on
Object ($+0.4$) but larger on Goal ($+3.4$) and Long ($+3.0$), where the
current observation alone may not determine the intended outcome or
subgoal order. The same pattern becomes more pronounced when the
demonstration is removed or replaced at evaluation
(Table~\ref{tab:abl_content}).

\begin{table}[t]
\centering
\scriptsize
% \small
\setlength{\tabcolsep}{2.7pt}
\begin{tabular}{l|ccc|ccc|ccc|ccc|ccc|ccc|ccc}
\toprule
\multirow{2}{*}{Task Suite} & \multicolumn{3}{c|}{OpenVLA-OFT} & \multicolumn{3}{c|}{LingBot-VLA} & \multicolumn{3}{c|}{Motus} & \multicolumn{3}{c|}{GR00T-N1.6} & \multicolumn{3}{c|}{Evo-Depth} & \multicolumn{3}{c|}{$\pi_{0.5}$} & \multicolumn{3}{c}{\textbf{StellaVLA}} \\
\cmidrule(lr){2-4}\cmidrule(lr){5-7}\cmidrule(lr){8-10}\cmidrule(lr){11-13}\cmidrule(lr){14-16}\cmidrule(lr){17-19}\cmidrule(lr){20-22}
& L0 & L1 & L2 & L0 & L1 & L2 & L0 & L1 & L2 & L0 & L1 & L2 & L0 & L1 & L2 & L0 & L1 & L2 & L0 & L1 & L2 \\
\midrule
\rowcolor{gray!20}\multicolumn{22}{@{}l}{\emph{Safety}} \\
Static Obst. & 0.89 & 0.25 & 0.21 & 0.47 & 0.00 & 0.00 & 0.82 & 0.56 & 0.19 & 0.72 & 0.30 & 0.14 & 0.88 & 0.66 & 0.48 & 0.90 & 0.64 & 0.50 & \textbf{1.00} & \textbf{1.00} & \textbf{0.80} \\
Cautious Grasp & 0.67 & 0.32 & 0.00 & 0.21 & 0.00 & 0.00 & 0.76 & 0.07 & \textbf{0.19} & 0.16 & 0.02 & 0.00 & 0.78 & 0.24 & 0.00 & 0.38 & 0.16 & 0.00 & \textbf{0.80} & \textbf{0.50} & 0.10 \\
Hazard Avoid. & 0.42 & 0.00 & 0.09 & 0.16 & 0.00 & 0.00 & 0.43 & 0.24 & 0.16 & \textbf{0.64} & 0.04 & 0.10 & 0.40 & 0.00 & 0.14 & 0.62 & 0.40 & 0.28 & 0.62 & \textbf{0.92} & \textbf{0.70} \\
State Pres. & \textbf{0.99} & 0.75 & 0.38 & 0.54 & 0.00 & 0.00 & 0.85 & 0.47 & 0.49 & 0.66 & 0.50 & 0.38 & 0.88 & 0.66 & 0.56 & 0.80 & 0.80 & \textbf{0.74} & 0.92 & \textbf{0.90} & 0.62 \\
Dynamic Obst. & 0.75 & 0.55 & 0.05 & 0.40 & 0.00 & 0.00 & 0.43 & 0.35 & \textbf{0.26} & 0.74 & 0.50 & 0.02 & \textbf{0.82} & 0.60 & 0.06 & 0.54 & \textbf{0.70} & \textbf{0.26} & 0.80 & 0.60 & 0.02 \\
\midrule
\rowcolor{gray!20}\multicolumn{22}{@{}l}{\emph{Distractor}} \\
Static Distr. & 0.99 & 0.13 & 0.15 & 0.93 & 0.15 & 0.11 & 0.75 & 0.19 & 0.03 & 0.46 & 0.32 & 0.06 & 0.94 & 0.20 & \textbf{0.24} & 0.90 & 0.04 & 0.14 & \textbf{1.00} & \textbf{0.56} & 0.20 \\
Dynamic Distr. & 0.90 & 0.56 & 0.39 & 0.88 & 0.61 & 0.17 & 0.73 & 0.60 & 0.33 & 0.70 & \textbf{0.72} & 0.18 & 0.86 & 0.60 & 0.32 & 0.88 & \textbf{0.72} & 0.56 & \textbf{1.00} & \textbf{0.72} & \textbf{0.60} \\
\midrule
\rowcolor{gray!20}\multicolumn{22}{@{}l}{\emph{Extrapolation}} \\
Prep. Comb. & 0.54 & 0.09 & 0.00 & 0.46 & 0.05 & \textbf{0.01} & 0.13 & 0.00 & 0.00 & 0.48 & 0.00 & 0.00 & 0.66 & 0.00 & 0.00 & 0.54 & 0.04 & 0.00 & \textbf{0.70} & \textbf{0.14} & 0.00 \\
Task Workflows & 0.53 & 0.03 & 0.13 & 0.37 & 0.05 & 0.11 & 0.32 & 0.00 & 0.02 & 0.42 & 0.00 & 0.00 & 0.32 & 0.00 & 0.00 & 0.52 & 0.22 & 0.26 & \textbf{0.64} & \textbf{0.66} & \textbf{0.84} \\
Unseen Obj. & 0.61 & 0.39 & 0.19 & 0.34 & 0.32 & 0.15 & 0.59 & 0.55 & 0.13 & 0.26 & 0.18 & 0.16 & 0.78 & 0.52 & 0.04 & 0.64 & 0.50 & 0.08 & \textbf{0.80} & \textbf{0.80} & \textbf{0.80} \\
\midrule
\rowcolor{gray!20}\multicolumn{22}{@{}l}{\emph{Long Horizon}} \\
Long Horizon & 0.80 & 0.00 & 0.00 & 0.82 & 0.03 & 0.00 & 0.64 & \textbf{0.05} & \textbf{0.03} & 0.29 & 0.02 & 0.00 & 0.93 & 0.00 & 0.00 & 0.87 & 0.00 & 0.00 & \textbf{1.00} & 0.02 & 0.00 \\
\midrule
\textit{Mean} & 0.74 & 0.28 & 0.14 & 0.51 & 0.11 & 0.05 & 0.59 & 0.28 & 0.17 & 0.50 & 0.24 & 0.09 & 0.75 & 0.32 & 0.17 & 0.69 & 0.38 & 0.26 & \textbf{0.84} & \textbf{0.62} & \textbf{0.43} \\
\midrule
\textit{Overall} & \multicolumn{3}{c|}{0.39} & \multicolumn{3}{c|}{0.22} & \multicolumn{3}{c|}{0.34} & \multicolumn{3}{c|}{0.28} & \multicolumn{3}{c|}{0.41} & \multicolumn{3}{c|}{0.44} & \multicolumn{3}{c}{\textbf{0.63}} \\
\bottomrule
\end{tabular}
\caption{Success rate on VLA-Arena across 3 difficulty
levels: L0 (in-distribution), L1 (intermediate generalization), and L2 (hardest). Rows are the $11$ task suites grouped into four categories (Safety, Distractor, Extrapolation, Long Horizon). Scores are fractions in $[0,1]$. Baseline results are from the VLA-Arena leaderboard~\citep{vlaarena}. Best per column in \textbf{bold}.}
\label{tab:vla_arena}
\end{table}

\subsubsection{Task-Level Generalization on VLA-Arena.}

VLA-Arena~\citep{vlaarena} evaluates task-level generalization across
11 suites in four categories: Safety, Distractor, Extrapolation, and
Long Horizon. Each suite contains three difficulty levels, from L0
(in-distribution) to L2 (hardest), while training uses L0 data only. We
follow the official protocol and compare against OpenVLA-OFT
\citep{openvlaoft}, LingBot-VLA~\citep{lingbotvla}, Motus~\citep{motus},
GR00T-N1.6~\citep{groot}, Evo-Depth~\citep{evodepth}, and
$\pi_{0.5}$~\citep{pi05}. At test time, StellaVLA receives one structured demonstration of the target task without any parameter update.
As shown in Table~\ref{tab:vla_arena}, StellaVLA achieves the best mean
success rate at all three levels: $0.84$, $0.62$, and $0.43$ on L0, L1,
and L2, respectively. Its overall score is $0.63$, compared with $0.44$ for
the strongest baseline, $\pi_{0.5}$. The margin over $\pi_{0.5}$ grows from
$0.15$ at L0 to $0.24$ at L1 and remains $0.17$ at L2, where no parameter
update is performed and only the target-task demonstration is supplied.

The per-suite results further delimit this gain. StellaVLA maintains $0.80$
on Unseen Objects across all three levels, consistent with transferring a
procedure when the object referent changes. Task Workflows rises from
$0.64$ at L0 to $0.84$ at L2, suggesting that context can become more useful
as the test workflow departs from training. Long Horizon remains near zero
at L1/L2 for every method, including ours: a fixed prefix specifies the
procedure but cannot re-plan after execution drift.

\subsubsection{Zero-Shot Robustness on LIBERO-Plus.}

LIBERO-Plus~\citep{liberoplus} perturbs LIBERO tasks along seven axes:
viewpoint, robot state, sensor noise, object layout, background, lighting,
and language. We evaluate the LIBERO-trained checkpoint without retraining
and compare with the zero-shot baselines reported in~\citep{liberoplus}.
StellaVLA reaches $85.1\%$ on average (Table~\ref{tab:libero_plus}),
outperforming StarVLA-OFT by $10.1$ points. The largest gains occur under
camera viewpoint ($+23.5$), sensor noise ($+19.7$), robot initial state
($+14.7$), and language perturbations ($+8.3$), where the observation or
instruction changes while the task procedure remains intact.

The smaller gains are also informative. Background and lighting are already
near saturation for both models, leaving little room for improvement. Under
object-layout changes, StellaVLA gains only $0.1$ point because the
demonstration's scene-specific spatial relation no longer matches the
current scene. Structured context therefore transfers task procedure more
reliably than a fixed spatial trajectory.

\begin{table}[t]
\centering
\small
\begin{tabular}{lccccccccc}
\toprule
Method & Orig. & Cam. & Robot & Noise & Layout & Backg. & Light & Lang. & Avg. \\
\midrule
OpenVLA        & 76.5 & 1.1  & 4.1  & 19.3 & 31.6 & 25.3 & 4.4  & 26.8 & 16.0 \\
OpenVLA-OFT    & 97.1 & 59.7 & 37.2 & 76.7 & 77.1 & 92.4 & 85.8 & 81.5 & 71.4 \\
$\pi_0$        & 94.2 & 15.8 & 6.6  & 79.4 & 70.4 & 78.5 & 79.6 & 61.0 & 53.8 \\
$\pi_0$-FAST   & 85.5 & 66.4 & 24.8 & 75.8 & 70.3 & 67.7 & 73.0 & 63.3 & 62.5 \\
Nora           & 87.9 & 4.0  & 41.1 & 17.6 & 63.9 & 50.5 & 31.0 & 67.0 & 38.7 \\
WorldVLA       & 79.1 & 0.3  & 30.2 & 12.2 & 39.4 & 14.5 & 29.4 & 44.2 & 24.3 \\
UniVLA         & 95.2 & 4.3  & 50.3 & 25.3 & 34.3 & 80.0 & 59.1 & 71.8 & 44.0 \\
RIPT-VLA       & 97.5 & 58.3 & 36.7 & 73.8 & 76.5 & 90.4 & 87.9 & 80.1 & 70.4 \\
StarVLA-OFT    & 96.6 & 47.0 & 60.1 & 73.1 & 79.2 & \textbf{95.3} & \textbf{96.3} & 87.0 & 75.0 \\
StellaVLA & \textbf{98.8} & \textbf{70.5} & \textbf{74.8} & \textbf{92.8} & \textbf{79.3} & 95.2 & 95.7 & \textbf{95.3} & \textbf{85.1} \\
\bottomrule
\end{tabular}
\caption{Zero-shot robustness on LIBERO-Plus~\citep{liberoplus}; all models
are trained on standard LIBERO and tested on the perturbed tasks without
retraining. \emph{Avg.}\ is the \emph{task-count-weighted} mean over the
seven perturbation categories, excluding \emph{Orig.} Best per column in
\textbf{bold}.}
\label{tab:libero_plus}
\end{table}

\subsection{Real-World and Cross-Embodiment Evaluation}

\noindent\textbf{Platform, Tasks, and Data.}
We evaluate on a $6$-DOF AgileX Piper arm with third-person and wrist-mounted
RGB cameras. The benchmark contains four tabletop tasks: precision
pen-to-cup placement, carrot-to-bowl pick-and-place, placing blocks in a
drawer and closing it, and stacking three bowls. We collect $125$
teleoperated robot episodes ($71{,}702$ frames), covering precision,
articulated-object, and multi-stage manipulation.

\noindent\textbf{Cross-Source Demonstration Pool.}
The retrieval pool combines the robot episodes with $26$ human-hand takes
recorded in XR and $26$ frame-aligned trajectories retargeted to the Piper.
All three sources are converted into the same structured representation.
The 2D grounding point is aligned across the human wrist and robot flange,
and off-embodiment frames are used only as retrieved context: all executable
action targets remain from the real robot.

\noindent\textbf{Protocol and Baselines.}
We evaluate four in-distribution tasks, two OOD-L1 object-attribute
perturbations, and one unseen OOD-L2 drawer task, with $10$ rollouts per
cell. Object placements are re-randomized and matched across methods.
StarVLA-OFT is the matched control for the complete StellaVLA design, while
the separately pretrained $\pi_{0.5}$~\citep{pi05}, fine-tuned on the same
episodes, provides an external reference. Appendix~B specifies the control
stack, perturbations, success criteria, and progress metric.

\begin{figure}[t]
\centering
\includegraphics[width=\linewidth]{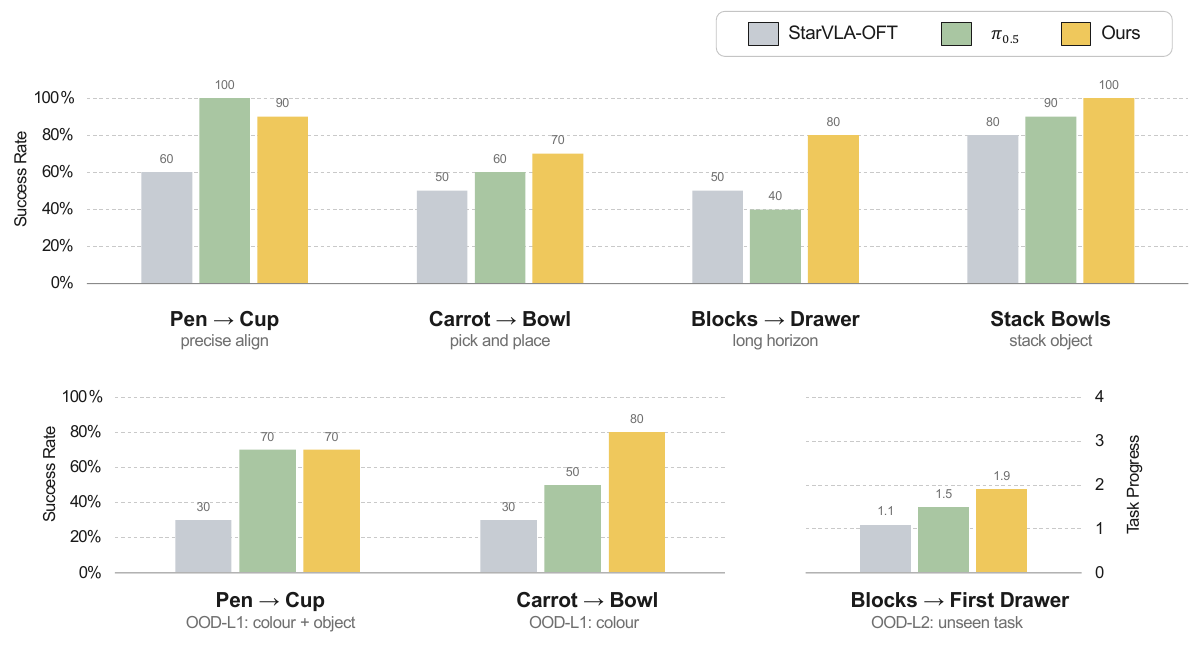}
\caption{Real-robot evaluation on the AgileX Piper. \emph{Top}: in-distribution success rate on the four tasks.
\emph{Bottom left}: \emph{OOD-L1}, which perturbs object attributes and is comparable within a task rather than across tasks. \emph{Bottom right}:
\emph{OOD-L2}, an unseen task placing the blocks in the first drawer.
Scenes are shown in Figure~\ref{fig:real_scene}.}
\label{fig:real_result}
\end{figure}

\begin{wraptable}{r}{0.38\columnwidth}
\centering
\vskip -0.1in
\small
\setlength{\tabcolsep}{6pt}
\begin{tabular}{lcc}
\toprule
Demo pair & ID & OOD \\
\midrule
$D(\text{real},\text{none})$  & 0.0041 & 0.0045 \\
$D(\text{human},\text{none})$ & 0.0041 & 0.0044 \\
$D(\text{xr},\text{none})$    & 0.0041 & 0.0044 \\
\midrule
$D(\text{human},\text{real})$ & 0.0015 & 0.0014 \\
$D(\text{xr},\text{real})$    & 0.0015 & 0.0016 \\
$D(\text{human},\text{xr})$   & 0.0014 & 0.0014 \\
\bottomrule
\end{tabular}
\caption{Normalized action disagreement when only the demonstration source
is changed.}
\label{tab:demosrc}
\end{wraptable}
\noindent\textbf{Results.}
StellaVLA reaches $85.0\%$ success in distribution and $75.0\%$ on OOD-L1
(Figure~\ref{fig:real_result}). On the two tasks with an L1 variant, its
average decreases from $80.0\%$ in distribution to $75.0\%$, a $5$-point
drop; StarVLA-OFT and $\pi_{0.5}$ drop by $25$ and $20$ points on the same
tasks. This smaller paired degradation indicates greater robustness when
the object attributes change, although the StarVLA-OFT comparison measures
the demonstration and spatial-language supervision jointly.

No method completes the unseen OOD-L2 task. StellaVLA nevertheless reaches
an average progress score of $1.9$ out of four, compared with $1.5$ for
$\pi_{0.5}$ and $1.1$ for StarVLA-OFT. This partial progress suggests that
the retrieved procedure remains useful, but does not constitute zero-shot
task completion.

\begin{figure}[t]
\centering
\includegraphics[width=\linewidth]{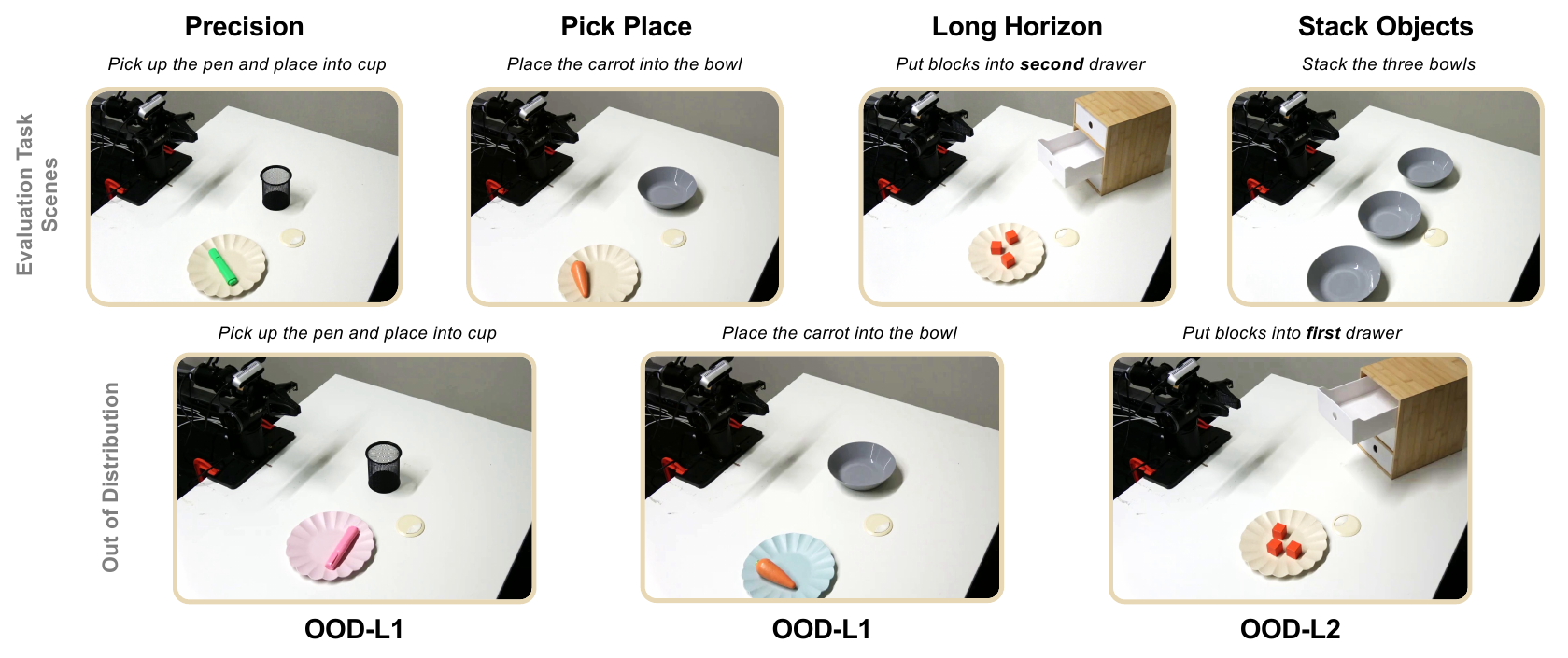}
\caption{\textbf{Real-robot evaluation scenes.} \emph{Top}: the four
in-distribution tasks. \emph{Bottom}: their out-of-distribution variants.
\emph{OOD-L1} substitutes both colour and object identity for
pen$\to$cup but colour alone for carrot$\to$bowl, which is why L1 is
comparable within a task rather than across tasks; \emph{OOD-L2} redirects
the blocks from the second drawer to the first, a task for which no
real-robot training episode exists.}
\label{fig:real_scene}
\end{figure}

\noindent\textbf{Cross-Source Consistency.}

The mixed-source rollouts above cannot isolate the effect of demonstration
provenance, because each cell samples from all three sources. We therefore
evaluate a fixed checkpoint on held-out teleoperated episodes while holding
the current observation fixed and changing only the retrieved source. We
measure the mean $L_1$ disagreement $D(S_1,S_2)$ between predicted action
chunks, normalized by the per-dimension standard deviation $\sigma$ of the
ground-truth actions. The paired human and XR-retargeted demonstrations
share the same trajectory and timing, so their comparison changes embodiment
appearance without changing the demonstrated behavior.

Across $1{,}780$ ID and $1{,}719$ OOD frames,
Table~\ref{tab:demosrc} shows that real-robot, human-hand, and XR-retargeted
context yield closely matched predictions. Source-to-source disagreement is
$0.0014$--$0.0016\sigma$, corresponding to $0.02^\circ$--$0.03^\circ$ of
joint angle and less than $0.05$\,mm of gripper width. This supports
consistency across the three structured demonstration sources.

This test is limited to single-step predictions. Removing the demonstration
changes the action by only $0.0041$--$0.0045\sigma$, and small differences
may still accumulate in closed-loop execution. The result therefore
establishes source consistency under matched observations, not closed-loop
invariance or the standalone contribution of context on hardware.

\subsection{Understanding Structured Demonstrations}
Unless stated otherwise, these analyses use the LIBERO checkpoint and report
the four-suite average (AVG) and, where available, the LIBERO-Plus average
(LP). Evaluation-time interventions change only the demonstration supplied
to a fixed checkpoint.

\begin{wraptable}{r}{0.42\columnwidth}
\centering
\vskip -0.1in
\small
\setlength{\tabcolsep}{4pt}
\begin{tabular}{lccccc}
\toprule
Demo & Sp. & Obj. & Goal & Long & AVG \\
\midrule
Correct & 99.6 & 99.0 & 99.6 & 96.8 & 98.8 \\
None    & 72.2 & 94.4 & 24.8 & 58.2 & 62.4 \\
Wrong   & 72.6 & 52.0 & \phantom{0}0.0 & 55.0 & 44.9 \\
\bottomrule
\end{tabular}
\caption{Changing demonstration content at evaluation.}
\label{tab:abl_content}
\vskip -0.1in
\end{wraptable}
\noindent\textbf{Causal Role of the Demonstration.}
Table~\ref{tab:abl_content} evaluates the same trained checkpoint with
the correct demonstration, no demonstration, and a demonstration from a
different task. Removing the demonstration reduces the average success
rate from $98.8$ to $62.4$, while providing a wrong-task demonstration
further lowers it to $44.9$. The latter result is particularly
informative: if the policy largely ignored the retrieved context, missing
and mismatched demonstrations would produce similar outcomes. Instead,
the additional degradation shows that StellaVLA actively uses the
demonstration to determine the intended task and adjusts its behavior
accordingly, even when the supplied context is misleading.

The per-suite results further clarify when this contextual specification
matters. On Goal, performance falls to $24.8$ without a demonstration and
to $0.0$ with a wrong one, since tasks may involve similar objects and
scenes but require different outcomes. The demonstration is therefore
critical for disambiguating the intended goal. By contrast, Spatial
performs similarly with no and wrong demonstrations ($72.2$ and $72.6$),
suggesting that the current observation still provides useful cues about
where the interaction should occur. These results provide direct evidence
that the retrieved demonstration serves as a task specification rather than
being ignored as auxiliary visual context.

\noindent\textbf{Spatial-Language Components.}
We next isolate the two movement fields in the structured
demonstration. Removing the 3D movement lowers AVG from $98.8$ to $97.8$,
whereas removing the 2D gripper path lowers it to $97.3$. Although the 2D
path is a projection of the same physical motion, it directly anchors that
motion in the current image. Its larger effect is consistent with the 2D
component providing visual grounding and the 3D component providing
workspace-level kinematic information. This experiment does not isolate the
semantic subtask label, for which no removal result is available.

\noindent\textbf{Demonstration Modality.}
Table~\ref{tab:abl_modality} examines which part of the demonstration
provides the useful context. At evaluation time, text-only demonstrations
nearly match the full image$+$text input, achieving $98.8/84.4$ compared
with $98.8/85.1$ on LIBERO AVG and LIBERO-Plus. In contrast, image-only
demonstrations decrease performance to $92.9/75.7$, with the largest loss
appearing on Long. These results indicate that most of the transferable
task information is captured by the structured language representation,
including the subgoal sequence and spatial movement, rather than by raw
demonstration frames alone. This also supports the use of demonstrations
across embodiments, since the structured description abstracts away much
of the source-specific visual appearance.

The training-time block of Table~\ref{tab:abl_modality}
reveals a complementary trade-off. Image-only demonstrations provide
slightly higher in-distribution performance than text-only demonstrations
($98.4$ vs.\ $97.3$), but generalize substantially worse on LIBERO-Plus
($78.7$ vs.\ $85.0$). A plausible explanation is that visual
demonstrations allow the policy to exploit appearance correspondence
between the demonstration and the current observation, which is effective
in distribution but becomes unreliable under visual perturbations.
Structured language removes much of this shortcut and instead encourages
the policy to rely on task-level and spatial structure that remains more
stable across distribution shifts.

\begin{wraptable}{r}{0.32\columnwidth}
\centering
\vskip -0.3in
\small
\setlength{\tabcolsep}{3pt}
\begin{tabular}{lcc}
\toprule
$\lambda$ & AVG & LP \\
\midrule
$0$   & 97.6 & \textbf{86.9} \\
$0.3$ & \textbf{98.8} & 85.1 \\
$0.6$ & 97.5 & 84.0 \\
$1.0$ & 97.2 & 81.9 \\
\bottomrule
\end{tabular}
\caption{Spatial-language loss weight.}
\label{tab:abl_langw}
\vskip -0.3in
\end{wraptable}
\begin{table}[t]
\centering
\small
\setlength{\tabcolsep}{10pt}
\begin{tabular}{lcccccc}
\toprule
Demo modality & Sp. & Obj. & Goal & Long & AVG & LP \\
\midrule
\rowcolor{gray!20}\multicolumn{7}{@{}l}{\emph{Changed at evaluation (fixed checkpoint)}} \\
Image$+$Text & 99.6 & 99.0 & 99.6 & 96.8 & \textbf{98.8} & \textbf{85.1} \\
Text-only    & 99.0 & 99.8 & 99.8 & 96.6 & \textbf{98.8} & 84.4 \\
Image-only   & 96.4 & 98.6 & 96.0 & 80.4 & 92.9 & 75.7 \\
\midrule
\rowcolor{gray!20}\multicolumn{7}{@{}l}{\emph{Changed at training}} \\
Text-only    & 98.6 & 98.6 & 99.2 & 92.8 & 97.3 & \textbf{85.0} \\
Image-only   & 99.4 & 99.4 & 98.8 & 95.8 & \textbf{98.4} & 78.7 \\
\bottomrule
\end{tabular}
\caption{Demonstration modality, changed at evaluation on a fixed
checkpoint (top) and at training (bottom). Image$+$Text is the default.
Best per column within each block in \textbf{bold}.}
\label{tab:abl_modality}
\end{table}

\noindent\textbf{Context Granularity.}
Performance changes by only $0.7$ point when the demonstration is reduced
from ten to three subgoal keyframes (Table~\ref{tab:abl_granularity} in
Appendix~C). This saturation suggests that the subgoal sequence, rather than
dense trajectory replay, carries the useful context.

\subsection{Spatial-Language Supervision and Deployment}

\noindent\textbf{Effect of Language-Loss Weight.}
Table~\ref{tab:abl_langw} varies the weight $\lambda$ of the
spatial-language objective. The in-distribution average follows a shallow
inverted-U and peaks at $\lambda=0.3$ ($98.8$), whereas LIBERO-Plus
robustness decreases from $86.9$ at $\lambda=0$ to $81.9$ at $\lambda=1.0$.
Spatial-language supervision therefore improves in-distribution precision
at the selected weight but does not monotonically improve OOD robustness;
context conditioning alone already provides a strong robustness signal.
One possible explanation is that a large language weight overemphasizes the
offline annotation schema. We use $\lambda=0.3$ as the best
in-distribution setting.

\begin{wraptable}{r}{0.42\columnwidth}
\centering
\small
\setlength{\tabcolsep}{6pt}
\begin{tabular}{lcc}
\toprule
Configuration & Cache & Latency \\
\midrule
No demo        & --- & 64 \\
Image$+$Text   & no  & 183 \\
Image$+$Text   & yes & 91 \\
Text-only      & --- & 109 \\
\midrule
\multicolumn{3}{l}{\emph{Paired language decoding}} \\
Action only    & yes & 88 \\
$+$ language   & yes & 3177 \\
\bottomrule
\end{tabular}
\caption{Inference latency (ms).}
\label{tab:abl_latency}
\vskip -0.1in
\end{wraptable}
\noindent\textbf{Inference Efficiency.}
Table~\ref{tab:abl_latency} separates the cost of context encoding from
language decoding. A forward pass without a demonstration takes $64$\,ms.
Adding an image$+$text demonstration raises this to $183$\,ms without a
cache, while caching the immutable prefix reduces the steady-state cost to
$91$\,ms. In a paired measurement, the action-only path takes $88$\,ms,
whereas decoding the $83$-token spatial-language output increases latency
to $3177$\,ms, about $36\times$ slower.

The $88$--$91$\,ms values measure the model-side path; the real-robot
pipeline, including observation and control overhead, runs at approximately
$205$\,ms per action chunk (Appendix~B). Thus, removing the
spatial-language expert avoids autoregressive decoding at deployment, while
prefix caching amortizes demonstration encoding across the rollout.

\section{Conclusion}

We presented \textbf{StellaVLA}, a framework that improves VLA
generalization by converting retrieved expert demonstrations into
structured in-context guidance. An automated offline pipeline extracts hierarchical semantic and kinematic rationales from heterogeneous demonstrations, while a parallel dual-training objective jointly learns continuous control and spatial-language reasoning. At inference, the spatial-language expert is removed and the fixed demonstration prefix is KV-cached, allowing the policy to benefit from structured reasoning without autoregressive language-decoding overhead. Experiments on LIBERO, VLA-Arena, LIBERO-Plus, and real-world manipulation show that StellaVLA preserves strong in-distribution performance while substantially improving robustness to task and environment shifts, including when demonstrations originate from different embodiments.

\section{Author List}

\noindent\textbf{Contributor:} 
Siyu Xu, Yunke Wang, Zijian Wang, Dihao Zhu, Chenghao Xia, Chengbin Du, Daochang Liu, Tao Huang, Chang Xu.

\clearpage
\bibliographystyle{plainnat}
\bibliography{references}

@inproceedings{hemotion,
  title={Motion Dynamics Learning for Few-Shot Embodied Adaptation},
  author={He, Sibo and Xie, Weiying and Li, Daixun and Zhong, Junhao and Tian, Jiayun and Wang, Yunke and Fang, Leyuan and He, Gang and Li, Yunsong},
  booktitle={Forty-third International Conference on Machine Learning}
}

@inproceedings{
feng2026see,
title={See What Matters: Differentiable Grid Sample Pruning for Generalizable Vision-Language-Action Model},
author={Yixu Feng and Zinan Zhao and Yanxiang Ma and Chenghao Xia and Chengbin Du and Yunke Wang and Chang Xu},
booktitle={Forty-third International Conference on Machine Learning},
year={2026}
}

@inproceedings{xie2026towards,
  title={Towards Efficient Embodied Reasoning: Mixture-of-Depth Compute Allocation for Vision-Language-Action Model},
  author={Xie, Weiying and Zeng, Qingchen and Meng, Zihan and Tian, Jiayun and He, Sibo and Yang, Danian and Du, Jie and Wang, Yunke and Li, Daixun and Wang, Hengyi and others},
  booktitle={Proceedings of the 32nd ACM SIGKDD Conference on Knowledge Discovery and Data Mining V. 2},
  pages={5732--5741},
  year={2026}
}

@article{qwen3vl,
  title   = {{Qwen3-VL Technical Report}},
  author  = {Bai, Shuai and Cai, Yuxuan and Chen, Ruizhe and Chen, Keqin and Chen, Xionghui and others},
  journal = {arXiv preprint arXiv:2511.21631},
  year    = {2025}
}

@article{openvla,
  title   = {{OpenVLA: An Open-Source Vision-Language-Action Model}},
  author  = {Kim, Moo Jin and Pertsch, Karl and Karamcheti, Siddharth and Xiao, Ted and Balakrishna, Ashwin and Nair, Suraj and Rafailov, Rafael and Foster, Ethan and Lam, Grace and Sanketi, Pannag and Vuong, Quan and Kollar, Thomas and Burchfiel, Benjamin and Tedrake, Russ and Sadigh, Dorsa and Levine, Sergey and Liang, Percy and Finn, Chelsea},
  journal = {arXiv preprint arXiv:2406.09246},
  year    = {2024}
}

@article{openvlaoft,
  title   = {{Fine-Tuning Vision-Language-Action Models: Optimizing Speed and Success}},
  author  = {Kim, Moo Jin and Finn, Chelsea and Liang, Percy},
  journal = {arXiv preprint arXiv:2502.19645},
  year    = {2025}
}

@article{pi0,
  title   = {{$\pi_0$: A Vision-Language-Action Flow Model for General Robot Control}},
  author  = {Black, Kevin and Brown, Noah and Driess, Danny and Esmail, Adnan and Equi, Michael and Finn, Chelsea and Fusai, Niccolo and Groom, Lachy and Hausman, Karol and Ichter, Brian and Jakubczak, Szymon and Jones, Tim and Ke, Liyiming and Levine, Sergey and Li-Bell, Adrian and Mothukuri, Mohith and Nair, Suraj and Pertsch, Karl and Shi, Lucy Xiaoyang and Tanner, James and Vuong, Quan and Walling, Anna and Wang, Haohuan and Zhilinsky, Ury},
  journal = {arXiv preprint arXiv:2410.24164},
  year    = {2024}
}

@article{pi0fast,
  title   = {{FAST: Efficient Action Tokenization for Vision-Language-Action Models}},
  author  = {Pertsch, Karl and Stachowicz, Kyle and Ichter, Brian and Driess, Danny and Nair, Suraj and Vuong, Quan and Mees, Oier and Finn, Chelsea and Levine, Sergey},
  journal = {arXiv preprint arXiv:2501.09747},
  year    = {2025}
}

@article{pi05,
  title   = {{$\pi_{0.5}$: A Vision-Language-Action Model with Open-World Generalization}},
  author  = {{Physical Intelligence} and Black, Kevin and Brown, Noah and Darpinian, James and Dhabalia, Karan and Driess, Danny and Esmail, Adnan and Equi, Michael and Finn, Chelsea and Fusai, Niccolo and Galliker, Manuel Y. and Ghosh, Dibya and Groom, Lachy and Hausman, Karol and Ichter, Brian and Jakubczak, Szymon and Jones, Tim and Ke, Liyiming and LeBlanc, Devin and Levine, Sergey and Li-Bell, Adrian and Mothukuri, Mohith and Nair, Suraj and Pertsch, Karl and Ren, Allen Z. and Shi, Lucy Xiaoyang and Smith, Laura and Springenberg, Jost Tobias and Stachowicz, Kyle and Tanner, James and Vuong, Quan and Walke, Homer and Walling, Anna and Wang, Haohuan and Yu, Lili and Zhilinsky, Ury},
  journal = {arXiv preprint arXiv:2504.16054},
  year    = {2025}
}

@article{groot,
  title   = {{GR00T N1: An Open Foundation Model for Generalist Humanoid Robots}},
  author  = {{NVIDIA} and Bjorck, Johan and Casta{\~n}eda, Fernando and Cherniadev, Nikita and Da, Xingye and Ding, Runyu and Fan, Linxi Jim and Fang, Yu and Fox, Dieter and Hu, Fengyuan and Huang, Spencer and Jang, Joel and Jiang, Zhenyu and Kautz, Jan and Kundalia, Kaushil and Lao, Lawrence and Li, Zhiqi and Lin, Zongyu and Lin, Kevin and Liu, Guilin and Llontop, Edith and Magne, Loic and Mandlekar, Ajay and Narayan, Avnish and Nasiriany, Soroush and Reed, Scott and Tan, You Liang and Wang, Guanzhi and Wang, Zu and Wang, Jing and Wang, Qi and Xiang, Jiannan and Xie, Yuqi and Xu, Yinzhen and Xu, Zhenjia and Ye, Seonghyeon and Yu, Zhiding and Zhang, Ao and Zhang, Hao and Zhao, Yizhou and Zheng, Ruijie and Zhu, Yuke},
  journal = {arXiv preprint arXiv:2503.14734},
  year    = {2025}
}

@article{univla,
  title   = {{UniVLA: Learning to Act Anywhere with Task-Centric Latent Actions}},
  author  = {Bu, Qingwen and Yang, Yanting and Cai, Jisong and Gao, Shenyuan and Ren, Guanghui and Yao, Maoqing and Luo, Ping and Li, Hongyang},
  journal = {arXiv preprint arXiv:2505.06111},
  year    = {2025}
}

@article{lapa,
  title   = {{Latent Action Pretraining from Videos}},
  author  = {Ye, Seonghyeon and Jang, Joel and Jeon, Byeongguk and Joo, Sejune and Yang, Jianwei and Peng, Baolin and Mandlekar, Ajay and Tan, Reuben and Chao, Yu-Wei and Lin, Bill Yuchen and Liden, Lars and Lee, Kimin and Gao, Jianfeng and Zettlemoyer, Luke and Fox, Dieter and Seo, Minjoon},
  journal = {arXiv preprint arXiv:2410.11758},
  year    = {2024}
}

@article{pi07,
  title   = {{$\pi_{0.7}$: A Steerable Generalist Robotic Foundation Model with Emergent Capabilities}},
  author  = {{Physical Intelligence}},
  journal = {arXiv preprint arXiv:2604.15483},
  year    = {2026}
}

@article{qwenrobotmanip,
  title   = {{Qwen-RobotManip Technical Report: Alignment Unlocks Scale for Robotic Manipulation Foundation Models}},
  author  = {{Qwen Team}},
  journal = {arXiv preprint arXiv:2606.17846},
  year    = {2026}
}

@article{nora,
  title   = {{NORA: A Small Open-Sourced Generalist Vision Language Action Model for Embodied Tasks}},
  author  = {Hung, Chia-Yu and Sun, Qi and Hong, Pengfei and Zadeh, Amir and Li, Chuan and Tan, U.-Xuan and Majumder, Navonil and Poria, Soujanya},
  journal = {arXiv preprint arXiv:2504.19854},
  year    = {2025}
}

@article{smolvla,
  title   = {{SmolVLA: A Vision-Language-Action Model for Affordable and Efficient Robotics}},
  author  = {Shukor, Mustafa and Aubakirova, Dana and Capuano, Francesco and Kooijmans, Pepijn and Palma, Steven and Zouitine, Adil and Aractingi, Michel and Pascal, Caroline and Russi, Martino and Marafioti, Andres and Alibert, Simon and Cord, Matthieu and Wolf, Thomas and Cadene, Remi},
  journal = {arXiv preprint arXiv:2506.01844},
  year    = {2025}
}

@article{starvla,
  title   = {{StarVLA: A Lego-like Codebase for Vision-Language-Action Model Developing}},
  author  = {{StarVLA Community}},
  journal = {arXiv preprint arXiv:2604.05014},
  year    = {2026}
}

@article{evodepth,
  title   = {{Evo-Depth: A Lightweight Depth-Enhanced Vision-Language-Action Model}},
  author  = {Lin, Tao and Du, Yuxin and Liu, Jiting and Zhu, Nuobei and Li, Yunhe and Fu, Yuqian and Chen, Yinxinyu and Cai, Hongyi and Ye, Zewei and Cheng, Bing and Ye, Kai and Mao, Yiran and Zhong, Yilei and Dong, MingKang and Yan, Junchi and Li, Gen and Zhao, Bo},
  journal = {arXiv preprint arXiv:2605.14950},
  year    = {2026}
}

@article{lingbotvla,
  title   = {{From Foundation to Application: Improving VLA Models in Practice}},
  author  = {Wu, Wei and Wang, Fangjing and Lu, Fan and Sun, He and Liu, Shi and Wang, Yunnan and Yan, Yibin and Wang, Yong and Ma, Shuailei and Wang, Xinyang and Liu, Yibin and Yang, Shuai and Zhou, Tianxiang and Zhang, Kejia and Zhou, Lei and Su, Cheng and Xue, Nan and Tan, Bin and Zhang, Han and Zhang, Youchao and Liao, Fei and Zhu, Xing and Shen, Yujun and Zheng, Kecheng},
  journal = {arXiv preprint arXiv:2607.06403},
  year    = {2026}
}

@article{motus,
  title   = {{Motus: A Unified Latent Action World Model}},
  author  = {Bi, Hongzhe and Tan, Hengkai and Xie, Shenghao and Wang, Zeyuan and Huang, Shuhe and Liu, Haitian and Zhao, Ruowen and Feng, Yao and Xiang, Chendong and Rong, Yinze and Zhao, Hongyan and Liu, Hanyu and Su, Zhizhong and Ma, Lei and Su, Hang and Zhu, Jun},
  journal = {arXiv preprint arXiv:2512.13030},
  year    = {2025}
}

@article{cotvla,
  title   = {{CoT-VLA: Visual Chain-of-Thought Reasoning for Vision-Language-Action Models}},
  author  = {Zhao, Qingqing and Lu, Yao and Kim, Moo Jin and Fu, Zipeng and Zhang, Zhuoyang and Wu, Yecheng and Li, Zhaoshuo and Ma, Qianli and Han, Song and Finn, Chelsea and Handa, Ankur and Liu, Ming-Yu and Xiang, Donglai and Wetzstein, Gordon and Lin, Tsung-Yi},
  journal = {arXiv preprint arXiv:2503.22020},
  year    = {2025}
}

@article{huang2026thinkact,
  title={Thinkact: Vision-language-action reasoning via reinforced visual latent planning},
  author={Huang, Chi-Pin and Wu, Yueh-Hua and Chen, Min-Hung and Wang, Frank and Yang, Fu-En},
  journal={Advances in Neural Information Processing Systems},
  volume={38},
  pages={82782--82802},
  year={2026}
}

@article{dreamvla,
  title   = {{DreamVLA: A Vision-Language-Action Model Dreamed with Comprehensive World Knowledge}},
  author  = {Zhang, Wenyao and Liu, Hongsi and Qi, Zekun and Wang, Yunnan and Yu, Xinqiang and Zhang, Jiazhao and Dong, Runpei and He, Jiawei and Lu, Fan and Wang, He and Zhang, Zhizheng and Yi, Li and Zeng, Wenjun and Jin, Xin},
  journal = {arXiv preprint arXiv:2507.04447},
  year    = {2025}
}

@article{memoryvla,
  title   = {{MemoryVLA: Perceptual-Cognitive Memory in Vision-Language-Action Models for Robotic Manipulation}},
  author  = {Shi, Hao and Xie, Bin and Liu, Yingfei and Sun, Lin and Liu, Fengrong and Wang, Tiancai and Zhou, Erjin and Fan, Haoqiang and Zhang, Xiangyu and Huang, Gao},
  journal = {arXiv preprint arXiv:2508.19236},
  year    = {2026}
}

@article{cogvla,
  title   = {{CogVLA: Cognition-Aligned Vision-Language-Action Model via Instruction-Driven Routing \& Sparsification}},
  author  = {Li, Wei and Zhang, Renshan and Shao, Rui and He, Jie and Nie, Liqiang},
  journal = {arXiv preprint arXiv:2508.21046},
  year    = {2025}
}

@inproceedings{avavla,
  title     = {{AVA-VLA: Improving Vision-Language-Action Models with Active Visual Attention}},
  author    = {Xiao, Lei and Li, Jifeng and Gao, Juntao and Ye, Feiyang and Jin, Yan and Qian, Jingjing and Zhang, Jing and Wu, Yong and Yu, Xiaoyuan},
  booktitle = {Proceedings of the IEEE/CVF Conference on Computer Vision and Pattern Recognition (CVPR)},
  pages     = {13453--13463},
  year      = {2026}
}

@inproceedings{retrievalvla,
  title     = {{Retrieval-VLA: Training-Free In-Context Adaptation for Vision-Language-Action Models}},
  author    = {Zhang, Yue and Wang, Rui and Lin, Jiehong and Wang, Zhongrui and Qi, Xiaojuan},
  booktitle = {Proceedings of the IEEE/CVF Conference on Computer Vision and Pattern Recognition (CVPR)},
  pages     = {1358--1367},
  year      = {2026}
}

@article{vivla,
  title   = {{See Once, Then Act: Vision-Language-Action Model with Task Learning from One-Shot Video Demonstrations}},
  author  = {Chen, Guangyan and Wang, Meiling and Shao, Qi and Zhou, Zichen and Mao, Weixin and Cui, Te and Zhu, Minzhao and Deng, Yinan and Yang, Luojie and Zhang, Zhanqi and Yang, Yi and Chen, Hua and Yue, Yufeng},
  journal = {arXiv preprint arXiv:2512.07582},
  year    = {2025}
}

@article{libero,
  title   = {{LIBERO: Benchmarking Knowledge Transfer for Lifelong Robot Learning}},
  author  = {Liu, Bo and Zhu, Yifeng and Gao, Chongkai and Feng, Yihao and Liu, Qiang and Zhu, Yuke and Stone, Peter},
  journal = {arXiv preprint arXiv:2306.03310},
  year    = {2023}
}

@article{liberoplus,
  title   = {{LIBERO-Plus: In-depth Robustness Analysis of Vision-Language-Action Models}},
  author  = {Fei, Senyu and Wang, Siyin and Shi, Junhao and Dai, Zihao and Cai, Jikun and Qian, Pengfang and Ji, Li and He, Xinzhe and Zhang, Shiduo and Fei, Zhaoye and Fu, Jinlan and Gong, Jingjing and Qiu, Xipeng},
  journal = {arXiv preprint arXiv:2510.13626},
  year    = {2025}
}

@article{vlaarena,
  title   = {{VLA-Arena: An Open-Source Framework for Benchmarking Vision-Language-Action Models}},
  author  = {Zhang, Borong and Li, Jiahao and Shen, Jiachen and Cai, Yishuai and Zhang, Yuhao and Chen, Yuanpei and Dai, Juntao and Ji, Jiaming and Yang, Yaodong},
  journal = {arXiv preprint arXiv:2512.22539},
  year    = {2025}
}

@article{xu2026vla,
  title={Vla-cache: Efficient vision-language-action manipulation via adaptive token caching},
  author={Xu, Siyu and Wang, Yunke and Xia, Chenghao and Zhu, Dihao and Huang, Tao and Xu, Chang},
  journal={Advances in Neural Information Processing Systems},
  volume={38},
  pages={164448--164473},
  year={2026}
}

@article{lap,
  title   = {{LAP: Language-Action Pre-Training Enables Zero-Shot Cross-Embodiment Transfer}},
  author  = {Zha, Lihan and Hancock, Asher J. and Zhang, Mingtong and Yin, Tenny and Huang, Yixuan and Shah, Dhruv and Ren, Allen Z. and Majumdar, Anirudha},
  journal = {arXiv preprint arXiv:2602.10556},
  year    = {2026}
}

@article{la4vla,
  title   = {{LA4VLA: Learning to Act without Seeing via Language-Action Pretraining}},
  author  = {Lin, Tao and Du, Yuxin and Mao, Yiran and Ye, Zewei and Zhong, Yilei and Cheng, Bing and Wang, Yiming and Liu, Jiting and Tian, Yang and Yan, Junchi and Wu, Feiran and Meng, Zenan and Wei, Hu and Fu, Yuqian and Li, Gen and Zhao, Bo},
  journal = {arXiv preprint arXiv:2606.27295},
  year    = {2026}
}

@article{actionsaslang,
  title   = {{Actions as Language: Fine-Tuning VLMs into VLAs Without Catastrophic Forgetting}},
  author  = {Hancock, Asher J. and Wu, Xindi and Zha, Lihan and Russakovsky, Olga and Majumdar, Anirudha},
  journal = {arXiv preprint arXiv:2509.22195},
  year    = {2025}
}

@article{steerablevla,
  title   = {{Steerable Vision-Language-Action Policies for Embodied Reasoning and Hierarchical Control}},
  author  = {Chen, William and Bhatia, Jagdeep Singh and Glossop, Catherine and Mathihalli, Nikhil and Doshi, Ria and Tang, Andy and Driess, Danny and Pertsch, Karl and Levine, Sergey},
  journal = {arXiv preprint arXiv:2602.13193},
  year    = {2026}
}

@inproceedings{zhang2026retrieval,
  title={Retrieval-VLA: Training-Free In-Context Adaptation for Vision-Language-Action Models},
  author={Zhang, Yue and Wang, Rui and Lin, Jiehong and Wang, Zhongrui and Qi, Xiaojuan},
  booktitle={Proceedings of the IEEE/CVF Conference on Computer Vision and Pattern Recognition},
  pages={1358--1367},
  year={2026}
}

@inproceedings{jangra,
  title={RA-VLA: Retrieval-Augmented VLA for Test-Time Adaptation},
  author={Jang, Sanghwan and Jeon, Minjin and Kim, Minsoo and Choi, Seong Jin and Kim, Dongha and Yu, Hwanjo},
  booktitle={Forty-third International Conference on Machine Learning},
  year={2026}
}

@article{lin2026la4vla,
  title={LA4VLA: Learning to Act without Seeing via Language-Action Pretraining},
  author={Lin, Tao and Du, Yuxin and Mao, Yiran and Ye, Zewei and Zhong, Yilei and Cheng, Bing and Wang, Yiming and Liu, Jiting and Tian, Yang and Yan, Junchi and others},
  journal={arXiv preprint arXiv:2606.27295},
  year={2026}
}

@article{zawalski2024robotic,
  title={Robotic control via embodied chain-of-thought reasoning},
  author={Zawalski, Micha{\l} and Chen, William and Pertsch, Karl and Mees, Oier and Finn, Chelsea and Levine, Sergey},
  journal={arXiv preprint arXiv:2407.08693},
  year={2024}
}

@article{lee2025molmoact,
  title={Molmoact: Action reasoning models that can reason in space},
  author={Lee, Jason and Duan, Jiafei and Fang, Haoquan and Deng, Yuquan and Liu, Shuo and Li, Boyang and Fang, Bohan and Zhang, Jieyu and Wang, Yi Ru and Lee, Sangho and others},
  journal={arXiv preprint arXiv:2508.07917},
  year={2025}
}

@inproceedings{acotvla,
  title     = {{ACoT-VLA: Action Chain-of-Thought for Vision-Language-Action Models}},
  author    = {Zhong, Linqing and Liu, Yi and Wei, Yifei and Xiong, Ziyu and Liu, Si and Ren, Guanghui},
  booktitle = {Proceedings of the IEEE/CVF Conference on Computer Vision and Pattern Recognition (CVPR)},
  pages     = {8152--8162},
  year      = {2026}
}

@article{behaviorprompt,
  title   = {{Behavior Prompting Policy: Demonstrations as Prompts for Manipulation}},
  author  = {Patel, Austin and Pekarek, Ben and Hernandez, Joel Enrique Castro and Song, Shuran},
  journal = {arXiv preprint arXiv:2606.30457},
  year    = {2026}
}

@article{bu2025univla,
  title={Univla: Learning to act anywhere with task-centric latent actions},
  author={Bu, Qingwen and Yang, Yanting and Cai, Jisong and Gao, Shenyuan and Ren, Guanghui and Yao, Maoqing and Luo, Ping and Li, Hongyang},
  journal={arXiv preprint arXiv:2505.06111},
  year={2025}
}

@article{robottt,
  title   = {{RoboTTT: Context Scaling for Robot Policies}},
  author  = {Jiang, Yunfan and Chebotar, Yevgen and Zheng, Ruijie and Hu, Fengyuan and Ge, Yunhao and Wu, Jimmy and Dai, Tianyuan and Reed, Scott and Fei-Fei, Li and Zhu, Yuke},
  journal = {arXiv preprint},
  year    = {2026}
}

@article{ttt,
  title   = {{Test-Time Training with Self-Supervision for Generalization under Distribution Shifts}},
  author  = {Sun, Yu and Wang, Xiaolong and Liu, Zhuang and Miller, John and Efros, Alexei A. and Hardt, Moritz},
  journal = {arXiv preprint arXiv:1909.13231},
  year    = {2020}
}

@article{tttvla,
  title   = {{TTT-VLA: Test-Time Latent Prompt Optimization for Vision-Language-Action Models}},
  author  = {Zhang, Wenbo and Li, Jianxiong and Yang, Shuai and Chen, Sijin and Liu, Jiajun and Liu, Lingqiao and Ma, Xiao},
  journal = {arXiv preprint arXiv:2606.03127},
  year    = {2026}
}

@article{wamttt,
  title   = {{WAM-TTT: Steering World-Action Models by Watching Human Play at Test Time}},
  author  = {Feng, Yusen and Han, Bingchen and Lyu, Jiangran and Liu, Kai and Zheng, Yixin and Wan, Yuxuan and Liu, Weiheng and Han, Sun and Li, Ruiqin and Zhang, Yulong and Liu, Fangfu and Shi, Xuesong and Liu, Libin and Wang, Yizhou and Zhang, Zhizheng and Wang, He},
  journal = {arXiv preprint arXiv:2607.06988},
  year    = {2026}
}

@article{hui2026seeing,
  title={Seeing Realism from Simulation: Efficient Video Transfer for Vision-Language-Action Data Augmentation},
  author={Hui, Chenyu and Huang, Xiaodi and Xu, Siyu and Wang, Yunke and You, Shan and Wang, Fei and Huang, Tao and Xu, Chang},
  journal={arXiv preprint arXiv:2605.02757},
  year={2026}
}

@misc{locoformer,
  title        = {{LocoFormer: Generalist Locomotion via Long-Context Adaptation}},
  key          = {LocoFormer},
  howpublished = {OpenReview},
  year         = {2025}
}

@article{dreamdojo,
  title   = {{DreamDojo: A Generalist Robot World Model from Large-Scale Human Videos}},
  author  = {Gao, Shenyuan and Liang, William and Zheng, Kaiyuan and Ye, Seonghyeon and Ma, Qianli and Zheng, Ruijie and Abbeel, Pieter and Zhu, Yuke and Jang, Joel and Fan, Linxi and others},
  journal = {arXiv preprint arXiv:2602.06949},
  year    = {2026}
}

@inproceedings{pei2026action,
  title={Action-aware dynamic pruning for efficient vision-language-action manipulation},
  author={Pei, Xiaohuan and Chen, Yuxing and Xu, Siyu and Wang, Yunke and Shi, Yuheng and Xu, Chang},
  booktitle={International Conference on Learning Representations},
  volume={2026},
  pages={10832--10851},
  year={2026}
}

@inproceedings{jang2026ravla,
  title     = {RA-VLA: Retrieval-Augmented VLA for Test-Time Adaptation},
  author    = {Sanghwan Jang and Minjin Jeon and Minsoo Kim and Seong Jin Choi and Dongha Kim and Hwanjo Yu},
  booktitle = {Proceedings of the 43rd International Conference on Machine Learning},
  year      = {2026},
  publisher = {PMLR},
  url       = {https://openreview.net/forum?id=ut6HebnnQe}
}

@inproceedings{xu2026affordance,
  title={Affordance field intervention: Enabling vlas to escape memory traps in robotic manipulation},
  author={Xu, Siyu and Wang, Zijian and Wang, Yunke and Xia, Chenghao and Huang, Tao and Xu, Chang},
  booktitle={Proceedings of the IEEE/CVF Conference on Computer Vision and Pattern Recognition},
  pages={37206--37215},
  year={2026}
}

@inproceedings{rth2024arxiv,
    title={RT-H: Action Hierarchies using Language},
    author={Suneel Belkhale and Tianli Ding and Ted Xiao and Pierre Sermanet and Quon Vuong and Jonathan Tompson and Yevgen Chebotar and Debidatta Dwibedi and Dorsa Sadigh},
    booktitle={https://arxiv.org/abs/2403.01823},
    year={2024}
}

@inproceedings{fu2025icrt,
  title={ICRT: In-Context Imitation Learning via Next-Token Prediction},
  author={Fu, Letian and Huang, Huang and Datta, Gaurav and Chen, Lawrence Yunliang and Panitch, William Chung-Ho and Liu, Fangchen and Li, Hui and Goldberg, Ken},
  booktitle={IEEE International Conference on Robotics and Automation (ICRA)},
  year={2025}
}

@article{sridhar2025ricl,
  title={RICL: Adding In-Context Adaptability to Pre-Trained Vision-Language-Action Models},
  author={Sridhar, Kaustubh and Dutta, Souradeep and Jayaraman, Dinesh and Lee, Insup},
  journal={arXiv preprint arXiv:2508.02062},
  year={2025}
}

@article{zhang2026revisiting,
  title={Revisiting Parameter Redundancy in Vision-Language-Action Models: Insights from VLM-to-VLA Adaptation},
  author={Zhang, Fengnian and Huang, Tao and Xu, Siyu and Jin, Zhong and Xu, Chang},
  journal={arXiv preprint arXiv:2606.31382},
  year={2026}
}

@article{fang2025intention,
  title={From intention to execution: Probing the generalization boundaries of vision-language-action models},
  author={Fang, Irving and Zhang, Juexiao and Tong, Shengbang and Feng, Chen},
  journal={arXiv preprint arXiv:2506.09930},
  year={2025}
}

@article{zhou2025libero,
  title={LIBERO-PRO: Towards Robust and Fair Evaluation of Vision-Language-Action Models Beyond Memorization},
  author={Zhou, Xueyang and Xu, Yangming and Tie, Guiyao and Chen, Yongchao and Zhang, Guowen and Chu, Duanfeng and Zhou, Pan and Sun, Lichao},
  journal={arXiv preprint arXiv:2510.03827},
  year={2025}
}

@article{wang2026x,
  title={X-OP: Cross-Morphology Whole-Body Teleoperation via MPC Retargeting},
  author={Wang, Jen-Wei and Kaingade, Sarthak and Tagliabue, Andrea and Morozovsky, Nicholas},
  journal={arXiv preprint arXiv:2606.07934},
  year={2026}
}

\clearpage
\appendix

\section{Implementation Details}

\noindent \textbf{Optimization.}
All models are fully fine-tuned from Qwen3-VL-4B-Instruct with AdamW
($\beta = 0.9/0.95$, weight decay $10^{-8}$), a backbone learning rate of
$5\times10^{-6}$ and an action-expert learning rate of $10^{-4}$ under a
cosine schedule ($500$ warmup steps), gradient clipping $1.0$, bf16 mixed
precision, and DeepSpeed ZeRO-2, for $30$k steps at a global batch size of
$128$.

\noindent \textbf{Regularization.}
Two per-sample dropouts regularize training. A 2D gripper-path dropout
($0.5$) keeps the 3D steering command usable when the pixel trace is
missing. A context-demonstration dropout matches demonstration
availability at deployment: $0.0$ when a same-task demonstration is always
retrievable (LIBERO, LIBERO-Plus) and $0.5$ otherwise (VLA-Arena, the real
robot).

\noindent \textbf{Retrieval.}
Demonstrations are retrieved by the language embedding of the task
instruction alone; no visual features enter the query. In simulation the
instruction set is closed, so this reduces to an exact task match, and we
condition on the single nearest demonstration ($M=1$) throughout. Training
uses leave-one-out retrieval, which excludes the target episode from the
pool.

\noindent \textbf{Action Expert and Spatial-Language Expert.}
A set of action-placeholder tokens is appended to the user turn, causally
preceding the language-action chain-of-thought. After a single backbone
forward, their final-layer hidden states pass through a two-block residual
MLP that regresses the action chunk under an $L_1$ objective. The compact
``subtask $+$ 3D/2D steering command'' chain-of-thought is retained from a
language-action formulation~\citep{lap} and supervised against
offline-generated annotations by cross-entropy; it is not decoded at
action inference. The steering command is the target $\Phi(A_t)$ of
Eq.~\ref{eq:chunktarget}, that is, the same action chunk the MLP regresses,
rendered by the verbaliser $\Phi$ over the chunk span rather than over a
whole segment; the subtask is the per-frame label $s_t$.

\noindent \textbf{Per-Platform Instantiation.}
In simulation the policy reads $256\times256$ images, predicts
end-effector deltas over an action chunk of length $8$, and the retrieved
demonstration carries up to ten subgoal keyframes. On the real robot it
reads $640\times480$ images, predicts a $7$-dimensional chunk of length
$16$ (six joint-position deltas relative to the episode start plus an
absolute gripper width, all $q_{99}$-normalized), and the demonstration
carries up to eight keyframes. The real-robot state is rendered in
language as the end-effector pose together with the joint angles.

\section{Real-World Setup}

\noindent \textbf{Control Stack.}
The AgileX Piper is driven over CAN at $30$\,Hz without a ROS stack. Each
$16$-step chunk spans $\approx\!0.53$\,s and is executed in a synchronous
receding-horizon loop with a horizon of $8$. Joint targets pass through a
One Euro filter and a $25^\circ$ per-step jump guard before reaching the
arm. Because the demonstration prefix is fixed for the duration of a
rollout, its key--value cache is computed once and reused, and steady-state
inference settles at $\approx\!205$\,ms per chunk, comfortably inside the
control budget.

\noindent \textbf{Data Collection.}
The $125$ teleoperated episodes ($71{,}702$ frames) were collected by
master--slave teleoperation across the four tasks: \emph{pick the pen and
put it in the cup}, \emph{pick the carrot and put it in the bowl},
\emph{put the blocks into the second drawer and close it}, and \emph{stack
the three bowls on top of each other}. They cover precision alignment,
free-space pick-and-place, a long-horizon sequence involving an articulated
object, and multi-stage stacking respectively.

\noindent \textbf{Cross-Source Pool Construction.}
Twenty-six human-hand takes recorded in XR yield two aligned datasets of
$26$ episodes and $10{,}996$ frames each: one preserving the operator's
bare hand, one showing a retargeted Piper arm executing the identical
trajectory. The two share episode structure and timing frame for frame, so
the appearance of the embodiment is the only variable separating them.
The grounding point of the 2D gripper path uses the operator's wrist for
the human take and the flange for the retargeted arm, both matched to the
real robot's convention, so that a rendered coordinate denotes the same
physical quantity regardless of provenance. The three sources are merged by
exact task-string match, retrieval excludes the current episode, and the
off-embodiment data enters \emph{only} the retrieval pool: no human or
retargeted frame ever supplies an action target, so the policy's motor
supervision remains entirely real-robot.

\noindent \textbf{Evaluation Conditions.}
Every reported cell is $10$ rollouts scored by a human operator against a
fixed success criterion, with object placements re-randomized per rollout
and held identical across methods. OOD-L1 severity is not calibrated across
tasks: it compounds a colour and an object substitution for pen$\to$cup but
changes only colour for carrot$\to$bowl, so L1 is comparable within a task
rather than across tasks. On OOD-L2 the progress score awards one point for
each of the three blocks placed and one for closing the drawer, out of
four. Single cells carry roughly $10$--$15$ points of noise at $10$
rollouts, so we base conclusions on the per-condition averages, which
aggregate $40$ rollouts in distribution and $20$ under L1, and on paired
degradations. Because only pen$\to$cup and carrot$\to$bowl have an L1
variant, the degradation quoted in the main text compares each policy
against its own score \emph{on those two tasks} ($80.0\%$ for StellaVLA,
$55.0\%$ for StarVLA-OFT, $80.0\%$ for $\pi_{0.5}$) rather than against the
four-task in-distribution average. On OOD-L2 the progress scores are $1.9$
for StellaVLA, $1.5$ for $\pi_{0.5}$ and $1.1$ for StarVLA-OFT; the task is
absent from real-robot training, so the human take is the only entry in the
pool holding a demonstration of it.

\begin{wraptable}{r}{0.32\columnwidth}
\centering
\vskip -0.1in
\small
\setlength{\tabcolsep}{6pt}
\begin{tabular}{lc}
\toprule
Subgoal frames & AVG \\
\midrule
$3$            & 98.1 \\
$5$            & 98.2 \\
$8$            & 98.3 \\
$10$ (default) & \textbf{98.8} \\
\bottomrule
\end{tabular}
\caption{Subgoal granularity. Best in \textbf{bold}.}
\label{tab:abl_granularity}
\vskip -0.2in
\end{wraptable}

\noindent \textbf{Baselines.}
The matched control is \emph{StarVLA-OFT}~\citep{starvla}, trained on the
same real-robot episodes with the same backbone, action expert and action
space, and differing only in receiving no demonstration and no auxiliary
spatial-language expert. We additionally fine-tune the openpi flow-matching
policy $\pi_{0.5}$~\citep{pi05} on the same episodes; as a separately
pretrained model family it places our absolute numbers on an external scale
rather than serving as a matched control.

\section{Additional Analysis}

\subsection{Demonstration Structure}
\noindent Subgoal granularity (Table~\ref{tab:abl_granularity}) is nearly flat:
three keyframes already reach $98.1$ AVG, within $0.7$ of the full ten
($98.8$). Denser sampling adds frames but no new sub-goals, and it is the
sub-goal decomposition the policy consumes, so returns saturate as soon as
the plan is complete rather than as soon as the trajectory is densely
covered. This is also what keeps the prefix short enough to cache.

\section{Discussion and Limitations}

\noindent \textbf{Language as a Cross-Embodiment Bridge.}
Under matched observations, real-robot, human-hand, and XR-retargeted
demonstrations produce closely matched action predictions
(Table~\ref{tab:demosrc}). The hardware results also use demonstrations
sampled from the merged three-source pool rather than from the robot source
alone. Together, these results indicate that the structured representation
reduces sensitivity to source-specific appearance and coordinates. They do
not establish closed-loop invariance across pinned sources, which requires a
larger source-controlled rollout study.

\noindent \textbf{Decoupling Reasoning from Latency.}
Autoregressive language generation is incompatible with the latency budget
of real-time control. StellaVLA instead trains the backbone under joint
action and spatial-language supervision, then removes the spatial-language
expert and caches the demonstration prefix at deployment. Control therefore
incurs no autoregressive language-decoding overhead
(Table~\ref{tab:abl_latency}).

\noindent \textbf{Limitations and Future Work.}
Despite its strong performance, our framework has several limitations.
\emph{Long-horizon compositionality:} while \textit{StellaVLA} substantially
improves task-level generalization and robustness, success rates on
extremely long-horizon, multi-stage tasks (such as the Long Horizon suite
in VLA-Arena) remain low for all evaluated models. The demonstration prefix
is fixed for the duration of an episode, so it cannot re-plan once
execution has drifted, and in-context conditioning alone therefore does not
resolve multi-step error accumulation.
\emph{Dependence on retrieval quality:} the effectiveness of our in-context
adaptation relies on retrieving a relevant expert demonstration. As the
wrong-demonstration ablation shows, conditioning on a mismatched
demonstration degrades performance below the no-demonstration baseline, so
more robust retrieval under visual and semantic domain shift is a critical
next step.
Finally, \emph{Coarse kinematic discretization:} our textualized kinematic rationale
relies on a quantized representation of spatial movements to fit within the
VLM's vocabulary. While this quantization suffices to shape the shared
representation, it may introduce discretization errors that limit the
precision of fine-grained manipulation. Continuous tokenization or hybrid
representation schemes could mitigate this.

\end{document}